\documentclass[12pt]{article}
\usepackage[utf8]{inputenc}
\usepackage{graphicx}
\usepackage{geometry}
\usepackage{amsmath}
\usepackage{hyperref}
\usepackage{placeins}
\begin{document}
\title{Reconceptualizing Smart Microscopy: From Data Collection to Knowledge Creation by Multi-Agent Integration}
\author{
  \hspace*{\fill} 
  \nobreakspace{} 
  P.S. Kesavan\textsuperscript{1,\dag}, Pontus Nordenfelt\textsuperscript{1,2,3,\dag} 
  \hspace*{\fill} \\[1em] 
  \small{
    \begin{minipage}{0.9\textwidth}
      \textsuperscript{1}Department of Clinical Sciences Lund, Infection Medicine, Faculty of Medicine, Lund University, Lund, Sweden \\
      \textsuperscript{2}Department of Laboratory Medicine, Clinical Microbiology, Skåne University Hospital Lund, Lund University, Lund, Sweden \\
      \textsuperscript{3}Science for Life Laboratory, Department of Clinical Sciences Lund, Lund University, Lund, Sweden \\[1em] 
      \textsuperscript{\dag}Correspondence to: \texttt{pskeshu@gmail.com} and \texttt{pontus.nordenfelt@med.lu.se}
    \end{minipage}
  }
}
\date{} 
\maketitle

\begin{abstract}
Smart microscopy represents a paradigm shift in biological imaging, moving from passive observation tools to active collaborators in scientific inquiry. Enabled by advances in automation, computational power, and artificial intelligence, these systems are now capable of adaptive decision-making and real-time experimental control. Here, we introduce a theoretical framework that reconceptualizes smart microscopy as a partner in scientific investigation. Central to our framework is the concept of the ”epistemic-empirical divide” in cellular investigation—the gap between what is observable (empirical domain) and what must be understood (epistemic domain). We propose six core design principles: epistemic-empirical awareness, hierarchical context integration, an evolution from detection to perception, adaptive measurement frameworks, narrative synthesis capabilities, and cross-contextual reasoning. Together, these principles guide a multi-agent architecture designed to align empirical observation with the goals of scientific understanding. Our framework provides a roadmap for building microscopy systems that go beyond automation to actively support hypothesis generation, insight discovery, and theory development, redefining the role of scientific instruments in the process of knowledge creation.
\end{abstract}

\section{Introduction}
The evolution of microscopy has been a cornerstone of biological research, enabling unprecedented insights into cellular structures and processes \cite{lichtman2005fluorescence, thorn2016quick}. From simple optical tools in the 17th century to automated systems generating vast quantities of standardized imaging data, microscopy has continually advanced the frontiers of cell biology \cite{kozubek2005image, gustafsson2005nonlinear, eliceiri2012biological}. Today, we stand at the cusp of another transformative transition—the emergence of smart microscopy systems that integrate artificial intelligence (AI) to enhance imaging capabilities and reshape cellular investigation \cite{scherf2015smart, mahecic2022event, carpenter2023smart, morgado2024rise}.

This transformation has been enabled by advances in automation, computational power, and AI, particularly the development of foundation models and real-time deep learning capabilities that enable systems to process and respond to complex biological data during experiments \cite{vaswani2017attention, ronneberger2015u, weigert2020star, chen2018allen, zhou2021informer}. These technologies have created unprecedented opportunities for microscopy systems to engage in complex decision-making and adaptive experimental control, enhancing imaging precision and efficiency \cite{chmielewski2015fast, royer2016adaptive, dunn2016brain, mahecic2022event, pylvanainen2023live}. Such capabilities are increasingly crucial for emerging fields like spatial transcriptomics and multiomics integration, where dynamic responses to complex biological presentations could dramatically enhance research outcomes \cite{skylaki2016challenges, misteli2020self, baysoy2023technological}.

However, smart microscopy represents more than just automated image acquisition or advanced computational analysis \cite{scherf2015smart,carpenter2023smart}. It embodies a paradigm shift from passive observation to active experimentation, with systems that dynamically engage with experimental processes \cite{mahecic2022event, andre2023data}. At its core, smart microscopy is distinguished by its capacity for real-time feedback and adaptation, enabling dynamic responses to emerging experimental conditions and research questions \cite{jackson2009intelligent, pylvanainen2023live, shroff2024live}.

Smart microscopy systems, integrating image acquisition with online analysis for data acquisition-image analysis feedback loops, have evolved significantly since their origins in the 1960s and 1970s, yet remain driven by technical advancements rather than conceptual advances in scientific inquiry \cite{wald1966contract, Herron1972AutomaticMF, Dew1974ANAM, morgado2024rise}. Leveraging modern hardware (optics, electronics, mechanics), software (AI, programming frameworks, user interfaces), and diverse applications, these systems have diversified into purpose-driven approaches:
\begin{itemize}
    \item Performance-driven approaches that optimize technical capabilities, such as throughput, real-time feedback, or event detection, through advanced hardware and AI \cite{conrad2011micropilot, mahecic2022event, fox2022enabling}.
    \item Application-driven approaches that target specific biological applications, such as live-cell imaging, super-resolution, or light-sheet microscopy \cite{pylvanainen2023live,  andre2023data, alvelid2022event, marin2024navigate, pedone2021cheetah}.
    \item Accessibility-driven approaches that enhance user-friendliness, generalizability, or inclusivity through intuitive interfaces or open-source platforms \cite{roos2024arkitekt, casas2021imswitch}.
\end{itemize}

Despite these advancements, smart microscopy remains conceptually limited to data acquisition-image analysis feedback loops, much like its early predecessors \cite{swedlow2012innovation, leonelli2019data, bechtel2010discovering}. While efficient for task-specific goals, such systems rarely engage with the dynamic workflows of biological knowledge gathering. To fundamentally transform experimentation and mechanize scientific insight generation, smart microscopy requires a theoretical framework that integrates it with the logic and workflows of scientific inquiry \cite{langley1987scientific, nickles2009scientific}. We argue that reconceptualizing smart microscopy as a collaborative partner, capable of advancing automated scientific discovery, will unlock its potential to revolutionize biological research \cite{eisenstein2020smart, king2009automation}.

Central to this reconceptualization is the recognition that scientific research navigates a fundamental divide between empirical observation and epistemic understanding \cite{popper2005logic, hacking1983representing}. Empirical aspects encompass what can be directly observed and measured—images, quantifications, and other data derived from microscopic investigation \cite{daston2021objectivity, chang2004inventing}. Epistemic aspects involve the creation of knowledge from these observations—developing theories, validating mechanisms, and generating insights that extend beyond direct measurement \cite{mitchell2009unsimple, rheinberger1997toward}. Smart microscopy must bridge this divide, integrating empirical data collection with epistemic goals to enhance the process of scientific discovery \cite{galison1997image, carlile2004transferring}.

This perspective presents a theoretical framework that addresses this challenge by positioning smart microscopy within the broader context of scientific inquiry \cite{kitano2016artificial, nielsen2020reinventing}. Our framework emphasizes that the ``smart'' in smart microscopy should represent the mechanization of researcher attributes—pattern recognition, contextual awareness, adaptive experimentation, and theoretical integration—within the specific constraints of microscopic investigation \cite{hoffman2018explaining, gil2019towards}. By establishing principles for navigating the interconnected domains of technical operation, experimental design, and knowledge creation, we aim to guide the development of systems that can meaningfully participate in the biological discovery process while addressing the unique epistemological challenges of cellular investigation \cite{sparkes2010towards}. To guide this transformation, we propose six core design principles—epistemic-empirical awareness, hierarchical context integration, a shift from detection to perception, adaptive measurement frameworks, narrative synthesis capabilities, and cross-contextual reasoning—that collectively enable smart microscopy to function as an active partner in scientific inquiry.

\section{The Epistemic-Empirical Divide in Cellular Investigation}

Cellular investigation presents unique challenges that highlight the fundamental tension between empirical observation and epistemic understanding in scientific research \cite{pawley2006fundamental, lambert2017navigating}. This divide is especially pronounced in microscopy, where researchers must build a comprehensive biological understanding from inherently fragmented and incomplete observations \cite{hacking1983representing, wolkenhauer2011complexity}.

\begin{figure}[htbp]
    \centering
    \includegraphics[width=0.8\textwidth]{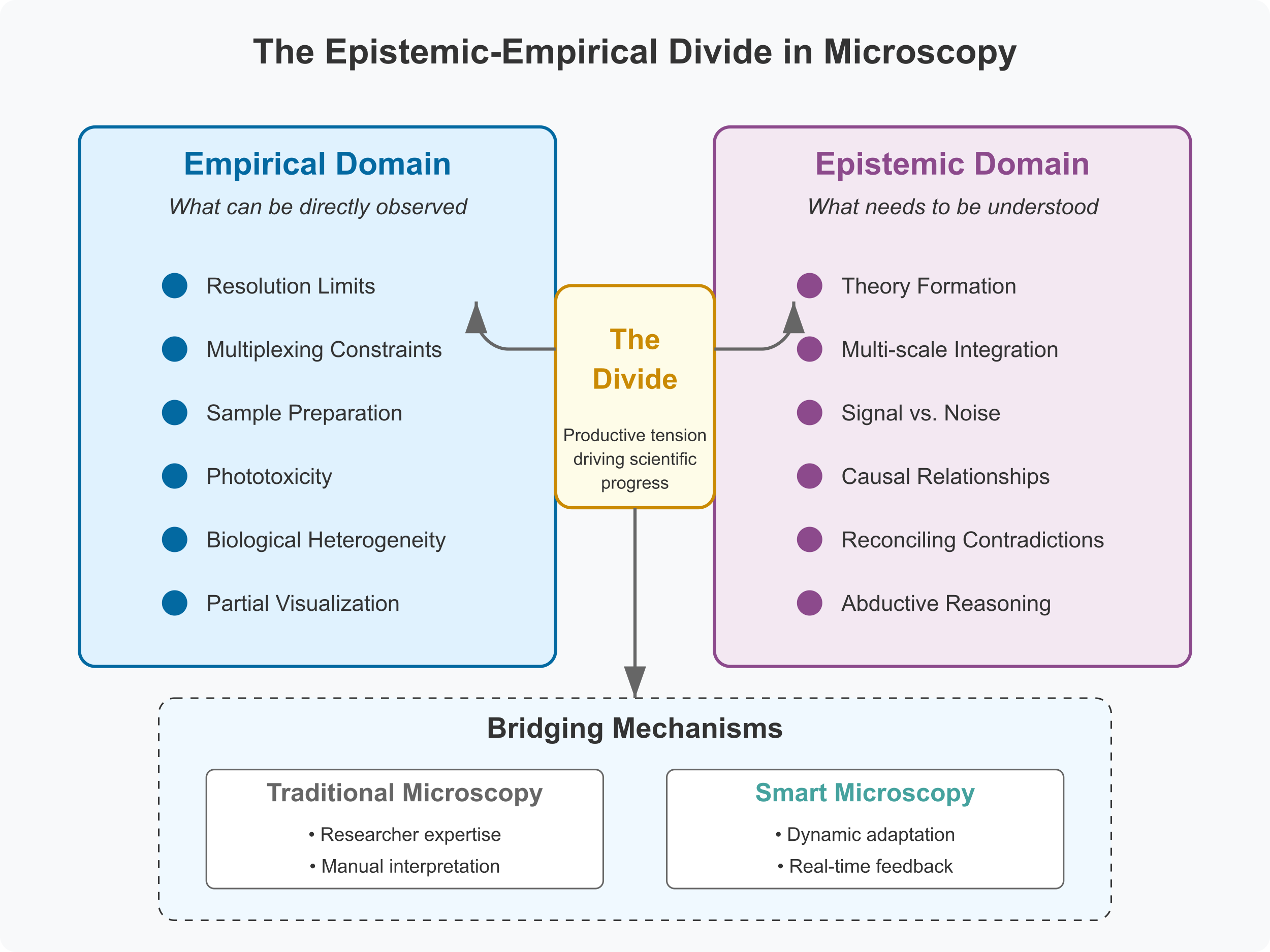}
    \caption{\textbf{The Epistemic-Empirical Divide in Microscopy.} The \textbf{empirical domain} encompasses what can be directly observed and measured through microscopic investigation: cellular structures, fluorescent signals, morphological features, temporal dynamics, and quantitative measurements. This domain is \textbf{constrained by} technical limitations including resolution boundaries, multiplexing capacity, photobleaching effects, sample preparation artifacts, and biological heterogeneity that limit complete observation of cellular reality. The \textbf{epistemic domain} encompasses what scientists need to understand: biological mechanisms, functional relationships, causal pathways, theoretical frameworks, and knowledge synthesis across multiple scales and contexts. This domain is \textbf{constrained by} the challenge of constructing coherent understanding from necessarily incomplete observations, integrating multi-scale phenomena, distinguishing biological variation from technical noise, and reconciling potentially contradictory evidence. The \textbf{epistemic-empirical divide} represents the fundamental gap between direct observation and theoretical understanding that creates productive tension driving scientific progress. Traditional microscopy approaches this divide sequentially (data collection followed by analysis), while smart microscopy enables dynamic navigation between empirical observation and epistemic understanding through real-time feedback, adaptive experimentation, and integrated interpretation that aligns data collection strategies with knowledge creation goals.}
    \label{fig:epistemic_empirical_divide}
\end{figure}

\subsection{Empirical Constraints in Microscopy}
The empirical domain of light microscopy is characterized by significant practical limitations that constrain what can be directly observed and measured. The most fundamental of these limitations is the physical boundary of resolution, which limits the visualization of structures below certain size thresholds \cite{waters2009accuracy}. Although super-resolution techniques have expanded these boundaries in light microscopy, they introduce their own constraints related to sample preparation, imaging speed, and phototoxicity \cite{cranfill2016quantitative}.

Fluorescence microscopy, the predominant method in cell biology, imposes additional empirical constraints due to its limited multiplexing capabilities \cite{schnell2012immunolabeling}. Most systems can visualize only a few markers simultaneously, forcing researchers to make critical decisions about which cellular components to observe during any given experiment. This partial visualization leads to an inherent incompleteness in empirical data—cellular structures and processes that are not labeled remain invisible, yet they may significantly influence the phenomena under investigation.

Sample preparation introduces another layer of empirical constraints. Live-cell imaging enables the observation of dynamic processes but limits resolution, marker options, and imaging duration due to phototoxicity concerns. Fixed-cell preparation allows for more detailed molecular visualization but sacrifices temporal information and introduces potential fixation artifacts. These trade-offs mean that empirical observations in microscopy are always partial approximations of cellular reality, filtered through technical limitations and methodological choices.

Perhaps the most challenging aspect is the inherent variability of biological samples \cite{altschuler2010cellular, Tanaka2016LiverRA}. Cells seeded on culture plates develop microenvironments influenced by subtle variations in density, neighboring cell interactions, and local substrate conditions. Even within a single experiment, regions may exhibit striking heterogeneity, creating distinct cellular contexts that influence behavior and response. This variability means empirical observations must always be interpreted with an awareness of contextual factors that may not be directly measurable.

\subsection{Epistemic Challenges in Cellular Understanding}
The epistemic domain concerns how researchers construct understanding from these inherently limited empirical observations. While this challenge exists across scientific disciplines, microscopy presents particularly pronounced manifestations of these epistemic challenges. First, there is the problem of theory formation from incomplete evidence  \cite{Bechtel2017UsingTH}. Researchers must develop coherent explanations for cellular phenomena using observations that capture only fragments of the complete biological reality. This requires sophisticated inferential processes to bridge the gaps between what can be directly observed and what must be understood.

Multi-scale integration presents another epistemic challenge \cite{Noble2012ATO, Lesne2013}. Cellular processes operate simultaneously across molecular, organellar, cellular, and tissue levels; however, most microscopy approaches focus on a limited range of these scales. Researchers must mentally integrate observations across different scales, often acquired through various imaging modalities with differing resolutions and marker sets. This integration necessitates conceptual frameworks that can bridge scale transitions and link molecular mechanisms to emergent cellular behaviors.

Particularly challenging is the epistemic task of distinguishing biological variation from technical variation \cite{Brigandt2010BeyondRA}. When cells exhibit heterogeneous responses, researchers must determine whether this indicates meaningful biological diversity or is merely the result of imaging, sample preparation, or analysis artifacts. This requires meta-knowledge about the technical limitations of methods and refined judgment concerning the distinction between signal and noise.

The need to reconcile potentially contradictory observations further complicates epistemic integration \cite{Sullivan2009TheMO, Gross2017}. Different imaging approaches may yield seemingly inconsistent results regarding the same biological process. Researchers must develop explanatory frameworks that can accommodate these contradictions, either by identifying contextual factors that clarify the differences or by refining theoretical models to incorporate apparently divergent observations.

\subsection{The Productive Tension}
The gap between empirical limitations and epistemic goals creates a productive tension that drives scientific progress in microscopy \cite{rheinberger1997toward}. This tension necessitates abductive reasoning—the inference to the best explanation—where researchers construct theoretical frameworks that account for limited and fragmentary observations \cite{Lipton1993InferenceTT}. The most valuable insights often emerge when researchers discover  creative ways to bridge this gap, developing new imaging approaches or analytical methods that transform empirical constraints into opportunities for discovery.

Historically, major advances in cell biology have often resulted from innovations that expand the empirical domain -- such as new visualization methods, increased multiplexing, and improved resolution -- or enhance epistemic integration through computational models, correlative techniques, and systems biology approaches. Smart microscopy represents the next frontier in this progression, with the potential to dynamically navigate the epistemic-empirical boundary by adapting data collection strategies based on emerging understanding and connecting observations to theoretical frameworks in real-time.

The challenge for smart microscopy, therefore, is not only to automate existing empirical methods but also to engage in this delicate navigation between what can be seen and what needs to be understood. This requires systems that acknowledge both the limitations of empirical observation and the goals of epistemic understanding—systems that can make informed decisions about what to measure, how to interpret measurements, and how to adapt investigative strategies based on emerging patterns and insights.
\FloatBarrier

\section{Scientific Insight in Microscopy}
Understanding how researchers generate insights from microscopic observation is essential for designing smart microscopy systems that can meaningfully participate in the scientific process. This section examines the cognitive, methodological, and social processes through which scientists transform visual data into biological understanding—processes that any truly ``smart'' system must engage with or enhance.

\subsection{Pattern Recognition and Comparative Analysis}
At the core of scientific insight in microscopy lies pattern recognition—the ability to identify meaningful structures, relationships, and changes within complex visual fields \cite{Kellman2009PerceptualLA}. Experienced researchers constantly compare current observations with mental models formed from past experiences, simultaneously detecting both confirmations and deviations from expected patterns \cite{Chi1981CategorizationAR}. This comparative process occurs across multiple levels of organization—from recognizing subcellular structures to evaluating experimental outcomes to assessing theoretical predictions.

The most valuable insights often emerge when observations challenge existing frameworks, creating a productive tension between what is seen and what was expected. For example, a researcher might notice unexpected protein localization patterns that contradict current models of cellular organization, prompting a reconsideration of fundamental assumptions about the system under study. These moments of recognition rely on the ability to perceive subtle deviations from expected patterns while filtering out technical artifacts and normal biological variation.

This pattern recognition process goes beyond simple feature detection \cite{Goodwin1994}. It entails understanding the biological significance of visual patterns and placing them within broader theoretical frameworks. A researcher doesn’t just identify a “punctate distribution” but sees it as potential vesicular trafficking, linking it to established models of membrane transport and considering its implications for cellular function.

\subsection{Contextual Integration}
Scientific insight emerges through contextual integration—the process of situating observations within interconnected experimental, theoretical, and conceptual frameworks \cite{Nersessian2008CreatingSC}. Researchers instinctively interpret microscopic images within multiple nested contexts: the specific experimental conditions, the broader research project, existing theoretical models, and the current state of the field. A cellular feature observed in isolation has limited meaning; the same feature interpreted within a disease model, signaling pathway, or developmental process provides much richer insights.

This contextual awareness allows researchers to move fluidly between immediate visual data and broader scientific questions \cite{Dunbar2005}. Their attention oscillates between a focused examination of specific features and a holistic assessment of patterns and relationships. For example, when examining images of cancer cells, a researcher might simultaneously consider specific morphological changes, their relationship to treatment conditions, potential molecular mechanisms driving these changes, and implications for therapeutic approaches.

The depth of contextual integration distinguishes novice observers from expert researchers \cite{Cetina1999-CETECH}. Novices often see isolated features and struggle to connect them to broader frameworks, whereas experts automatically place their observations within rich conceptual contexts that inform their interpretations and guide further investigations.

\subsection{Creative Exploration and Serendipity}
While scientific investigation is often portrayed as methodical and systematic, the generation of insights relies heavily on creative exploration and serendipitous discovery \cite{Nickles1980ScientificDL}. The most significant breakthroughs frequently arise at the boundaries of planned investigations—when researchers notice unexpected patterns, follow intuitive hunches, or make opportunistic observations that weren't part of the original experimental design.

These moments typically arise not from rigidly following protocols but from maintaining an open awareness while systematically exploring a phenomenon \cite{Bedessem2019ScientificAA}. Researchers develop what might be called "disciplined serendipity"—creating conditions where unexpected observations can occur while upholding the rigorous methods needed to distinguish meaningful patterns from artifacts or coincidences.

This balance between structured analysis and creative exploration represents one of the most challenging aspects of mechanizing scientific cognition. It requires a simultaneous commitment to methodological rigor and an openness to detecting patterns beyond predefined categories or measurements. Smart microscopy systems must navigate this balance by providing systematic data collection while supporting the identification and exploration of unexpected phenomena.

\subsection{The Social Dimension of Insight}

Scientific understanding arises not only from individual observation but also through the social processes of collective interpretation and validation \cite{Lynch1985ArtAA, Latour1986-LATLLT}. In laboratory meetings, conferences, and collaborative discussions, researchers refine their interpretations through dialogue with colleagues who offer different perspectives, expertise, and theoretical frameworks regarding the same visual data.

This social dimension of insight generation operates through specific communicative practices \cite{Star}. Researchers use specialized terminology, metaphorical language, and visual representations to convey their observations and interpretations. When a scientist describes membrane dynamics as "ruffling" or protein distribution as "scaffolding," they are using metaphorical language that encodes both descriptive information and theoretical perspectives. These linguistic and representational practices facilitate the collective construction of meaning from visual data.

The social nature of scientific understanding has profound implications for smart microscopy design. Systems must not only process images and generate measurements but also engage in the communicative practices through which scientific meaning emerges. This mandates capabilities for generating descriptions that connect to established biological language, creating visualizations that highlight meaningful patterns, and potentially participating in discussions about the importance of observations.

\subsection{The Epistemic Toolkit of Microscopy Researchers}
Researchers have developed sophisticated tools and practices to bridge the epistemic-empirical divide in microscopy. These include:
\begin{itemize}
    \item \textbf{Linguistic frameworks} that translate visual patterns into biologically meaningful concepts \cite{Leonelli2016DataCentricBA}. The specialized vocabulary of cell biology—terms like ``punctate,'' ``diffuse,'' ``polarized,'' or ``colocalized''—represents a sophisticated system for encoding visual information in ways that connect to theoretical understanding.
    \item \textbf{Visual representations} that formalize relationships between cellular components and processes \cite{Ankeny_Leonelli_2021}. Diagrams, models, and other visualizations serve as cognitive scaffolds that help researchers integrate fragmentary observations into coherent conceptual frameworks.
    \item \textbf{Mental modeling} capabilities that enable researchers to simulate cellular processes that extend beyond direct observation \cite{Merz2006EmbeddingDI}. Scientists mentally animate static images, visualize molecular interactions past resolution limits, and conceive three-dimensional structures from two-dimensional slices.

    \item \textbf{Methodological adaptations} that address the limitations of individual techniques. Researchers strategically combine complementary approaches—correlative microscopy, live-cell followed by fixed-cell imaging, or multiplexed marker sets across multiple samples—to overcome the episodic fragmentation inherent in microscopic observation.
    \item \textbf{Quantitative frameworks} that translate visual patterns into numerical representations. They serve as bridges between empirical observation and theoretical models, enabling researchers to transition between qualitative assessment and mathematical description.
\end{itemize}

These tools collectively enable researchers to construct coherent biological narratives from inherently incomplete observations. They exemplify sophisticated adaptations to the fundamental constraints of microscopic investigation—adaptations with which any smart microscopy system must interact to become a meaningful partner in scientific inquiry.

\section{Redefining Smart Microscopy}
The preceding analysis of the epistemic-empirical divide and the mechanics of scientific insight provides the foundation for reconceptualizing smart microscopy. Instead of simply enhancing automation or computational capabilities, we propose a fundamental shift in the conceptualization and design of these systems. This section delineates this redefinition and its implications for the development of next-generation microscopy systems.

\subsection{From Tool to Partner in Scientific Inquiry}

Traditional microscopy systems, even those with advanced automation capabilities, primarily function as tools—they execute predefined imaging protocols and analysis routines but remain fundamentally passive instruments in the research process \cite{king2009automation}. Smart microscopy represents a transformation of these systems from tools to partners in scientific inquiry—active participants in the iterative cycle of observation, interpretation, and experimental adaptation that characterizes biological investigation.

This shift requires transcending  the current paradigm in which intelligence is primarily applied to image analysis after data collection \cite{kitano2016artificial}. Instead, intelligence must be integrated throughout the research process—from experimental design and data acquisition to analysis, interpretation, and hypothesis generation. A truly smart microscope doesn’t merely automate existing workflows; it actively engages in the decision-making processes that guide scientific investigation.

This redefinition positions smart microscopy not as an endpoint but as a specialized instance of a broader category: artificial scientific inquiry \cite{bates2020reporting}. While adjacent fields like robotic scientists, digital twins, and AI-based scientific assistants share this general orientation, smart microscopy represents a distinct domain with unique challenges and requirements. It specifically focuses on the intersection between visual information and biological understanding, addressing the particular complexities of constructing knowledge from microscopic observation.

\subsection{Mimicking Researcher Attributes within Microscopic Investigation}

The "smart" in smart microscopy can be understood as the mechanization of researcher attributes—the cognitive, methodological, and social capabilities that enable scientists to generate insights from microscopic observation \cite{langley1987scientific}. These include:

\begin{itemize}
    \item \textbf{Pattern recognition and anomaly detection} capabilities that identify both expected structures and deviations from predicted patterns.
    \item \textbf{Contextual awareness} that situates observations within experimental conditions, research questions, and theoretical frameworks.
    \item \textbf{Adaptive experimentation} approaches that modify imaging parameters, field selection, or temporal sampling based on emerging observations.
    \item \textbf{Integrative reasoning} that connects observations across different scales, modalities, and time points into coherent biological narratives.
    \item \textbf{Communicative capabilities} that translate visual patterns into biologically meaningful descriptions and facilitate dialogue about their significance.
\end{itemize}

These capabilities represent a collaborative approach to scientific discovery rather than automation alone \cite{gil2014amplify}.

Importantly, these capabilities must be tailored to the unique context of microscopic investigation. Unlike general AI systems, smart microscopy must tackle the particular challenges of cellular observation, including issues with incomplete information, integration of scales, and the episodic nature of microscopic data collection.

This specialization indicates that smart microscopy systems require deep domain knowledge in cellular biology, experimental methods, and microscopy techniques \cite{walmsley2021artificial}. They must understand both the empirical constraints of microscopic observation—resolution limits, photobleaching concerns, and sample variability—as well as the epistemic frameworks that guide biological interpretation.

\subsection{Bridging the Epistemic-Empirical Divide}

The central promise of smart microscopy lies in its potential to bridge the epistemic-empirical divide that characterizes cellular investigation \cite{rheinberger2020epistemology}. Current approaches typically address only one side of this divide, either enhancing empirical data collection through improved automation and standardization or supporting epistemic processes through advanced analysis and visualization. Smart microscopy should aim to integrate these domains, aligning data collection strategies with knowledge creation goals and connecting experimental decisions to theoretical frameworks.

This integration requires bidirectional translation between empirical observations and epistemic constructs \cite{lakatos1978methodology}. Systems must translate research questions into suitable imaging strategies, convert visual patterns into biologically meaningful concepts, and link quantitative measurements to qualitative interpretations. At the same time, they must translate emerging understanding back into empirical approaches, identifying what additional observations are needed to validate hypotheses, resolve uncertainties, or explore unexpected phenomena.

By navigating this boundary, smart microscopy can address the fundamental challenge of cellular investigation: constructing a coherent understanding from necessarily fragmented observations \cite{schaffner1995discovery}. Systems can aid researchers in bridging episodic gaps, connecting observations across different samples, time points, and imaging modalities. They can support integration across scales, linking molecular mechanisms to cellular behaviors and tissue-level phenomena. Additionally, they can facilitate the identification of patterns and relationships that might otherwise remain obscured by the complexity and volume of microscopic data.

This redefinition has profound implications for the design, implementation, and evaluation of smart microscopy systems. Success cannot be measured merely by technical metrics such as throughput, resolution, or computational efficiency. Instead, these systems must be evaluated on their capacity to enhance the scientific inquiry process itself, to generate insights, identify meaningful patterns, suggest productive experiments, and ultimately accelerate the development of biological understanding.

\section{Design Principles for Next-Generation Smart Microscopy}
Building on our redefinition of smart microscopy as a partner in scientific inquiry, we propose six core design principles to guide the development of next-generation systems. These principles address the unique challenges of microscopic investigation while leveraging emerging capabilities in artificial intelligence, computer vision, and automated experimentation.

\subsection{Epistemic-Empirical Awareness}

Smart microscopy systems must simultaneously recognize empirical limitations and epistemic goals, acknowledging the constraints of what can be directly observed while orienting data collection toward what needs to be understood \cite{chang2004inventing}. This dual awareness should be explicit in system architecture and operation, featuring clear representations of both technical parameters (exposure times, resolution limits, signal-to-noise considerations) and research objectives (mechanism validation, phenotype characterization, structure-function relationships).

This awareness requires internal models of the relationship between empirical data and epistemic claims in biological research. Systems should maintain explicit representations of:
\begin{itemize}
    \item \textbf{Technical constraints} that limit empirical observation, including resolution boundaries, photobleaching concerns, multiplexing limitations, and sample preparation artifacts.
    \item \textbf{Inferential pathways} that connect empirical observations to biological conclusions, including both direct measurements and the interpretive frameworks used to derive meaning from these measurements.
    \item \textbf{Uncertainty landscapes} that map relative confidence in different aspects of current understanding, identifying areas where empirical evidence is strong versus domains where conclusions rely heavily on theoretical inference.
\end{itemize}

This balance between empirical precision and epistemic relevance reflects the inherent complexity of biological systems \cite{mitchell2009unsimple, daston2021objectivity}.

This principle allows systems to make informed decisions about resource allocation, determining when to invest in additional empirical data collection versus when to focus on alternative interpretations of existing data. It also promotes transparency in scientific communication, clarifying the connection between directly observed phenomena and inferences drawn from theoretical reasoning.

\subsection{Hierarchical Context Integration}
\begin{figure}[htbp]
    \centering
    \includegraphics[width=0.8\textwidth]{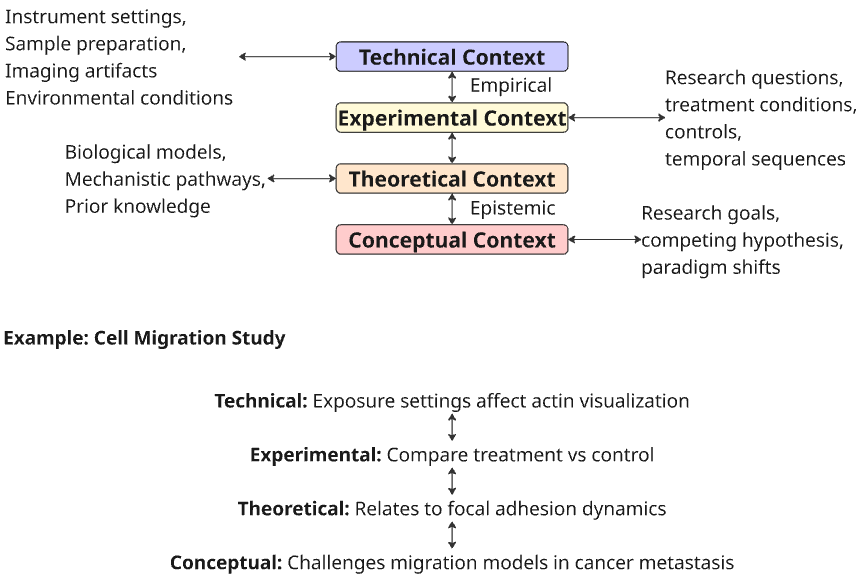}
\caption{\textbf{Hierarchical Context Integration in Smart Microscopy.} Smart microscopy systems must simultaneously navigate and integrate four interconnected levels of context to make contextually appropriate decisions. The \textbf{technical context} forms the empirical foundation, encompassing instrument settings, sample preparation protocols, imaging artifacts, and environmental conditions that influence observation quality. The \textbf{experimental context} bridges empirical observations with epistemic goals, including research questions, treatment conditions, control parameters, and temporal sequences that frame current observations within the overall experimental design. The \textbf{theoretical context} provides epistemic frameworks through biological models, mechanistic pathways, and prior knowledge that inform interpretation of visual data and connect observations to established cellular structures and processes. The \textbf{conceptual context} represents the highest level, encompassing overarching research goals, competing hypotheses, and potential paradigm shifts that drive investigation. Bidirectional arrows demonstrate how changes at one level propagate to others, creating a dynamic system where technical adjustments influence experimental comparisons, theoretical interpretations, and conceptual implications. The example demonstrates this integration in a cell migration study, showing how technical imaging decisions connect to experimental comparisons, theoretical mechanisms, and conceptual understanding of cancer metastasis.}
\label{fig:hierarchical_context_integration}
\end{figure}

Smart microscopy systems must operate simultaneously across multiple levels of context, from technical imaging parameters to broader scientific frameworks. These contexts form a natural hierarchy:
\begin{itemize}
    \item \textbf{Technical context} serves as the empirical foundation—instrument settings, sample preparation protocols, imaging artifacts, and environmental conditions that influence observation. This also includes an understanding of how factors like exposure settings, objective selection, or sample mounting affect image quality and interpretation.
    \item \textbf{Experimental context} serves as the bridge between empirical observation and epistemic goals, encompassing research questions, treatment conditions, control parameters, and temporal sequences that frame current observations. This includes understanding how current imaging relates to the overall experimental design and what comparisons or contrasts are scientifically meaningful.
    \item \textbf{Theoretical context} provides the epistemic framework—biological models, mechanistic pathways, and prior knowledge that inform the interpretation of visual data. This includes an awareness of how observations map to established cellular structures, processes, and relationships.
    \item \textbf{Conceptual context} represents the highest level—overarching research goals, competing hypotheses, and potential paradigm shifts that drive investigation. This includes understanding how current experiments contribute to broader scientific questions and identifying observations that would have significant implications for existing theories.
\end{itemize}

This multi-level awareness reflects the knowledge-level perspective in cognitive architectures \cite{newell1982knowledge}.

The key design challenge is maintaining dynamic relationships between these contexts \cite{anderson2002spanning, marr2010vision}. Changes at one level should propagate appropriately to others—a shift in exposure settings (technical) should be interpreted in terms of its impact on experimental comparisons, theoretical interpretations, and conceptual implications. Conversely, changes in research questions (conceptual) should drive suitable adjustments to experimental design, theoretical focus, and technical parameters.

This hierarchical integration enables smart microscopy systems to make contextually appropriate decisions regarding data collection, analysis, and interpretation. It also facilitates communication with researchers at the most relevant level of abstraction—from technical discussions about imaging parameters to conceptual dialogues about research implications.
\FloatBarrier

\subsection{Evolution from Detection to Perception}

\begin{figure}[htbp]
    \centering
    \includegraphics[width=\textwidth]{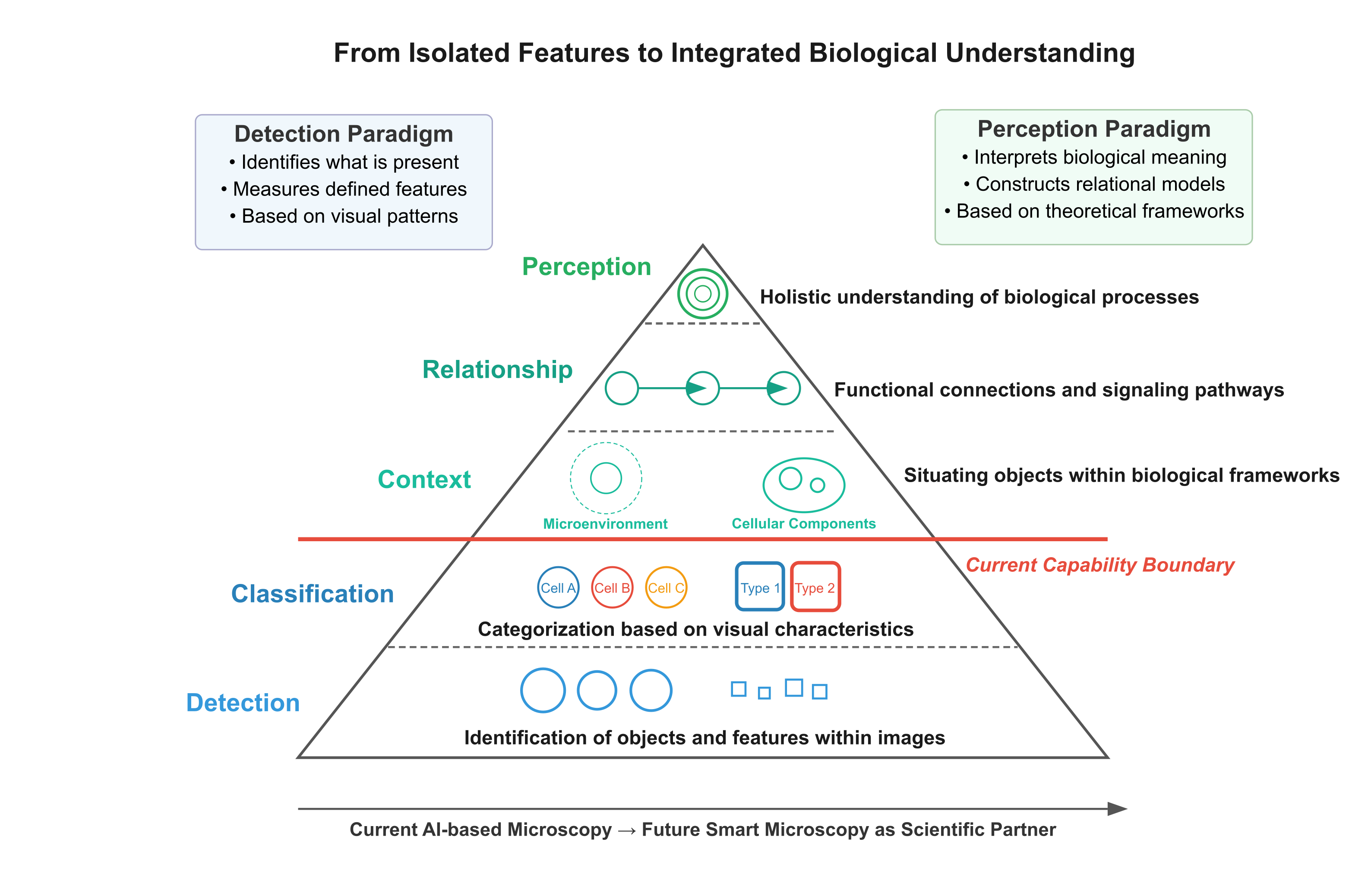}
\caption{\textbf{Evolution from Detection to Perception in Smart Microscopy.} This hierarchical model illustrates the critical transition from basic computer vision detection to advanced biological perception capabilities that smart microscopy systems must achieve. The pyramid represents five progressive layers: (1) \textbf{Detection} identifies basic objects and features within images; (2) \textbf{Classification} categorizes detected elements based on visual characteristics and morphological properties; (3) \textbf{Context} situates objects within biological frameworks, including both the cellular microenvironment (extracellular matrix, neighboring cells, local biochemical factors) and internal cellular organization (organelles, cytoskeletal structures, functional domains); (4) \textbf{Relationship} recognizes functional connections, signaling pathways, and causal associations between perceived elements; and (5) \textbf{Perception} achieves holistic understanding of biological processes that extends beyond visual features to include biological significance, temporal behavior, and theoretical relevance. The red line indicates the current capability boundary of existing AI-based microscopy systems, which excel at detection and classification but struggle with higher-level contextual understanding. Future smart microscopy systems must evolve toward true perception capabilities, representing a fundamental paradigm shift from identifying what is visually present to interpreting what is biologically happening within specific theoretical frameworks.}    \label{fig:detection-perception}
\end{figure}

Current computer vision approaches in microscopy excel at detection—identifying cells, nuclei, or specific morphological features within images \cite{vu2019methods} (Fig.~\ref{fig:detection-perception}). While valuable, detection alone does not constitute perception in the sense that researchers perceive microscopic specimens. True perception requires situating detected elements within meaningful relationships and biological contexts.

Next-generation smart microscopy must evolve from detection to perception through several capabilities \cite{meijering2016imagining}:
\begin{itemize}
    \item \textbf{Perception of objects} that extend beyond visual features to include biological significance, temporal behavior, and theoretical relevance. Rather than simply identifying a structure as a "vesicle," systems should recognize it as part of an endocytic pathway, potentially containing specific cargo, and participating in cellular adaptation to experimental conditions.
    \item \textbf{Relationship recognition} that identifies not just spatial proximity but functional associations, causal relationships, and theoretical connections between perceived elements. Systems should recognize that membrane protrusions relate to migration behavior, that nuclear translocation suggests signaling activation, or that morphological changes correlate with cell cycle progression.
    \item \textbf{Cross-modal integration} that combines information across different imaging approaches, fluorescent markers, or contrast methods to create a unified perceptual representations. Similar to how researchers mentally integrate DAPI-stained nuclei with phalloidin-labeled actin from separate images, systems should construct integrated perceptual models that transcend individual acquisition channels.
    \item \textbf{Temporal continuity} that maintains object identity and tracks relationships across time points despite cellular movement, division, or morphological changes. This requires sophisticated object permanence capabilities that persist through partial occlusion, temporary marker loss, or significant morphological transformation.
\end{itemize}

This evolution requires moving beyond training models to recognize patterns toward systems that can reason about what those patterns mean within specific biological contexts \cite{moen2019deep}. It represents a shift from ``what is visually present'' to ``what is biologically happening''—the same transition that researchers make when interpreting microscopic images.
\FloatBarrier

\subsection{Adaptive Measurement Frameworks}

Microscopy inherently involves measurement—converting visual observations into quantitative representations that can be analyzed, compared, and related to theoretical models \cite{chang2007scientific}. However, the choices made in measurement embed assumptions about which features are relevant, what constitutes meaningful variation, and how visual information should be reduced to numerical representations.

Smart microscopy systems should expose and adapt these measurement frameworks based on research context and emerging data patterns:
\begin{itemize}
    \item \textbf{Measurement transparency} makes explicit the choices underlying measurement approaches—why particular features are quantified, what thresholds define categories, and how complex visual information is distilled into specific metrics. This transparency enables both systems and researchers to critically evaluate measurement validity in various contexts.
    \item \textbf{Contextual adaptation} modifies measurement strategies based on experimental conditions, cell types, or research questions. Different cellular processes may require entirely different measurement frameworks—what is relevant for migration studies differs fundamentally from what matters in division analysis.
    \item \textbf{Multi-framework application} utilizes various complementary measurement approaches simultaneously, acknowledging that different quantification strategies can unveil distinct aspects of biological phenomena. Instead of adhering to a single measurement framework, systems should uphold multiple perspectives on the same visual data.
    \item \textbf{Measurement evolution} refines quantification approaches based on emerging patterns in the data or feedback from researchers, while maintaining rigorous controls against overfitting. When initial measurements fail to capture important variations or produce unexpected distributions, systems should suggest alternative quantification strategies that may better represent the biological phenomena, but must distinguish between discovering meaningful biological patterns and artificially finding structure in noise. This balance requires explicit mechanisms for validation, hypothesis testing, and maintaining scientific integrity in the measurement refinement process.
\end{itemize}

These approaches acknowledge that measurement is inherently a theory-laden activity that influences scientific understanding \cite{tal2015measurement, van2010scientific}. 

This principle moves beyond configuration wizards with predefined options to systems that can reason about measurement choices in the context of specific experimental goals and emerging observations. It recognizes that measurement is not a neutral process but an interpretive one that profoundly shapes which patterns can be discovered and what conclusions can be drawn.

\subsection{Narrative Synthesis Capabilities}
\begin{figure}[htbp]
    \centering
    \includegraphics[width=0.8\textwidth]{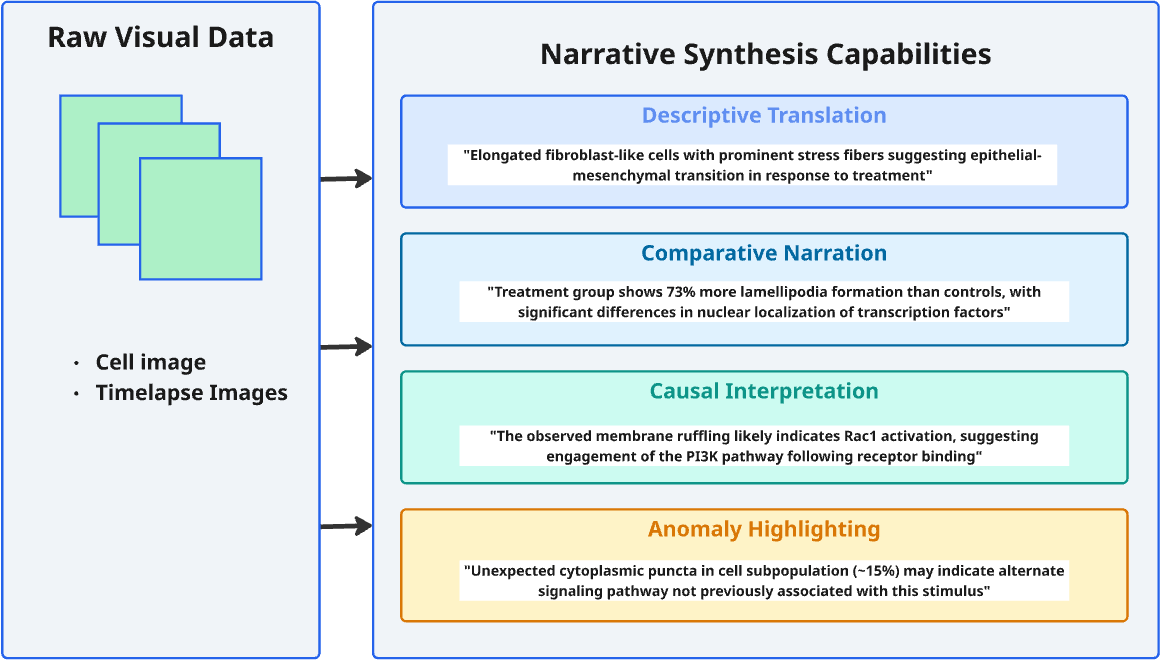}
\caption{\textbf{Narrative Synthesis Capabilities in Smart Microscopy.} This figure illustrates how smart microscopy systems transform raw visual data into increasingly sophisticated biological narratives through four progressive levels of interpretation. Using an example context of cell migration analysis, we demonstrate: (1) \textbf{Descriptive translation} converts visual patterns into biologically meaningful language, moving beyond simple feature detection (e.g., "elongated cells") to contextually rich descriptions (e.g., "fibroblast-like morphology suggesting epithelial-mesenchymal transition"); (2) \textbf{Comparative narration} highlights biologically significant similarities and differences across experimental conditions, time points, or cellular populations, focusing on variations that relate to research questions rather than all detectable differences; (3) \textbf{Causal interpretation} proposes potential mechanisms or explanations for observed phenomena through abductive reasoning, connecting visual observations to theoretical frameworks while acknowledging uncertainty; and (4) \textbf{Anomaly highlighting} identifies unexpected patterns, contradictions, or novel connections that might escape human attention, focusing on biologically significant deviations from expected patterns or theoretical predictions. This hierarchical approach enables smart microscopy systems to generate coherent biological stories that bridge empirical observation with epistemic understanding, providing structured frameworks that enhance rather than replace human interpretation in the scientific discovery process.}
\label{fig:narrative_synthesis}
\end{figure}

Perhaps the most significant gap in current smart microscopy is the inability to generate meaningful narratives from observations—the translation of visual patterns and measurements into coherent biological stories that explain cellular phenomena \cite{dahlstrom2014using} (Fig.~\ref{fig:narrative_synthesis}). Vision-language models offer promising capabilities for bridging this gap \cite{cho2021unifying}.

Narrative synthesis in smart microscopy should include:
\begin{itemize}
    \item \textbf{Descriptive translation} that converts visual patterns into biologically meaningful language. Beyond simple feature labeling, this involves characterizing complex phenomena in terms that connect to established biological concepts—describing not just "elongated cells" but "fibroblast-like morphology suggesting epithelial-mesenchymal transition".
    \item \textbf{Comparative narration} that highlights similarities and differences across experimental conditions, time points, or cellular populations. These comparisons should focus on biologically significant variations rather than all detectable differences, prioritizing changes that relate to research questions or suggest new hypotheses.
    \item \textbf{Causal interpretation} that proposes potential mechanisms or explanations for observed phenomena. While acknowledging uncertainty, systems should suggest plausible biological processes that could generate observed patterns—connecting visual observations to theoretical frameworks through abductive reasoning.
    \item \textbf{Anomaly highlighting} that identifies unexpected patterns, contradictions, or connections that might escape human attention. This "narrative anomaly detection" focuses not on technical artifacts but on biologically significant deviations from expected patterns or theoretical predictions.
\end{itemize}

These narrative capabilities connect data to meaning through techniques similar to those used in human scientific reasoning \cite{herman2017storytelling, morgan2017narrative}.

These narratives would not replace human interpretation but provide structured frameworks that accelerate and enhance the meaning-making process. They make explicit connections that might otherwise remain implicit, suggest interpretations that researchers might not immediately consider, and highlight patterns that could otherwise be overlooked in complex or high-dimensional data.
\FloatBarrier

\subsection{Cross-Contextual Reasoning}

Scientific progress often emerges through abductive reasoning—inferring the most likely explanation for observed phenomena by linking empirical observations to theoretical frameworks \cite{magnani2011abduction, douven2011abduction}. Smart microscopy systems should support this reasoning process through cross-contextual integration.

This capability includes:
\begin{itemize}
    \item \textbf{Gap identification} that recognizes discrepancies between observations and expectations according to current theory. Systems should highlight not only technical anomalies but also conceptually significant deviations that may indicate limitations in existing models or opportunities for theoretical refinement.
    \item \textbf{Alternative interpretation generation} that proposes multiple possible explanations for unexpected results. Instead of committing to a single explanation, systems should maintain and evaluate competing hypotheses that could account for observed patterns.
    \item \textbf{Experiment suggestion capabilities} that recommend follow-up investigations to differentiate between competing explanations or to fill gaps in current understanding. These suggestions should be concrete and specific, considering available techniques, reagents, and research priorities.
    \item \textbf{Knowledge integration} that connects new observations with existing research literature, identifying both supporting evidence and potential contradictions. This approach contextualizes current findings within a broader scientific understanding and highlights implications for established models.
\end{itemize}

This principle advances smart microscopy from data analysis to knowledge synthesis, transforming the task from merely describing observations to reasoning about their significance and determining what to investigate next. It represents the highest level of partnership in scientific inquiry, where systems actively engage in the iterative process of refining hypotheses and developing theories.

\section{Implementation Architecture}
Translating these design principles into functional systems requires an architecture that integrates multiple types of intelligence while managing the complexities of microscopic investigation. We propose a multi-agent framework that distributes scientific cognition across specialized components while maintaining cohesive operation across the epistemic-empirical divide.

\subsection{Multi-Agent Approach to Scientific Cognition}
\begin{figure}[htbp]
    \centering
    \includegraphics[width=0.8\textwidth]{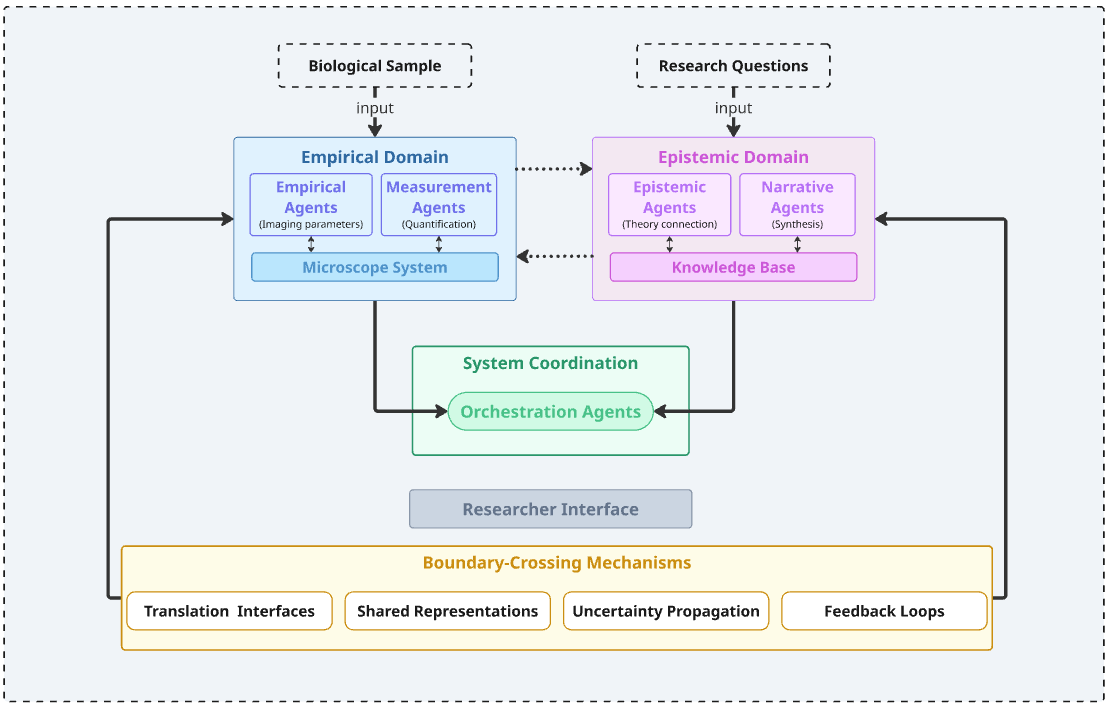}
\caption{\textbf{Multi-Agent Architecture for Smart Microscopy.} This diagram illustrates the proposed multi-agent framework that distributes scientific cognition across specialized components to enable effective navigation of the epistemic-empirical divide. The architecture operates across two primary domains: the \textbf{Empirical Domain} (left, blue) receives biological samples as input and manages data acquisition through Empirical Agents (controlling imaging parameters and optimizing acquisition settings) and Measurement Agents (applying quantitative frameworks and extracting meaningful metrics) that interface directly with the Microscope System. The \textbf{Epistemic Domain} (right, purple) processes research questions through Epistemic Agents (connecting observations to theoretical frameworks) and \textbf{Narrative Agents} (synthesizing information into coherent descriptions and explanations), supported by a comprehensive Knowledge Base containing biological models and prior research. \textbf{Orchestration Agents} (green) coordinate overall investigation strategy, balancing exploration with confirmation while maintaining alignment with research goals. Critical \textbf{Boundary-Crossing Mechanisms} (yellow) enable bidirectional information flow through Translation Interfaces, Shared Representations, Uncertainty Propagation, and Feedback Loops that connect empirical observations with epistemic understanding. The Researcher Interface facilitates human-AI collaboration throughout the system, allowing scientists to engage at appropriate levels of abstraction from technical parameter discussions to conceptual dialogues about research implications.}    \label{fig:multi_agent_architecture}
\end{figure}

The cognitive processes involved in microscopic investigation are too complex and diverse to be effectively implemented in a monolithic system \cite{wooldridge2009introduction}. Instead, we propose a multi-agent architecture where specialized agents collaborate to support various aspects of scientific inquiry:

\begin{itemize}
    \item \textbf{Empirical agents} focus on the technical domain of microscopy—controlling imaging parameters, optimizing acquisition settings, and processing raw visual data. These agents maintain detailed models of microscope capabilities, sample properties, and imaging physics to ensure high-quality data collection. They understand empirical trade-offs (resolution vs. speed, signal vs. phototoxicity) and can adaptively optimize these parameters based on higher-level goals.
    
    \item \textbf{Measurement agents} employ quantitative frameworks to extract meaningful metrics from visual data. These agents maintain multiple measurement approaches, propose appropriate quantification strategies for different biological questions, and adjust analytical methods based on emerging data patterns. They operate at the intersection of empirical observation and epistemic interpretation, translating visual information into quantitative representations that support scientific reasoning.

    \item \textbf{Epistemic agents} connect observations to theoretical frameworks and research questions. These agents maintain models of biological processes, experimental designs, and research objectives to interpret empirical data within broader scientific contexts. They can identify significant patterns, suggest potential mechanisms, and evaluate how new observations support or challenge existing theories.
    \item \textbf{Narrative agents} synthesize information across other agents to generate coherent descriptions, explanations, and visualizations. These agents translate technical details, measurements, and interpretations into language and representations that effectively communicate with researchers. They highlight key findings, contextualize observations, and produce reports at suitable levels of technical detail for different communication purposes.
    \item \textbf{Orchestration agents} coordinate the overall investigation strategy, balancing exploration with confirmation while maintaining alignment with research goals. These agents prioritize imaging targets, suggest experimental sequences, and allocate system resources based on scientific objectives. They stay aware of the overall research context and ensure that individual observations contribute to a broader understanding.
\end{itemize}

This distributed intelligence approach leverages specialized expertise while enabling collaborative problem-solving \cite{jennings2000agent, stone2000multiagent}.

This multi-agent approach allows specialization within domains while facilitating integration across the epistemic-empirical divide. Agents can develop deep expertise in their specific areas and communicate through well-defined interfaces that support collaborative scientific reasoning.

\subsection{Boundary-Crossing Mechanisms}

The most critical aspect of this architecture is implementing effective mechanisms for crossing boundaries between empirical and epistemic domains \cite{galison1997image}. These mechanisms facilitate the flow of information and decision-making across traditionally separate aspects of scientific investigation:

\begin{itemize}
    \item \textbf{Translation interfaces} convert information between domains—linking research questions to imaging parameters, visual features to biological concepts, and quantitative measurements to qualitative interpretations. These interfaces ensure that epistemic goals guide empirical collection strategies and that empirical observations deepen epistemic understanding.

    \item \textbf{Shared representations} serve as common ground between domains—models that simultaneously represent empirical details (cell boundaries, marker distributions) and epistemic concepts (cell types, functional states, process stages). These representations facilitate reasoning across domains while maintaining connections to both visual data and theoretical frameworks.
    \item \textbf{Uncertainty propagation mechanisms} explicitly track confidence levels across domain boundaries. Empirical uncertainty (measurement noise, detection confidence) must be accurately translated into epistemic uncertainty (confidence in interpretations, support for hypotheses). This transparency prevents unwarranted certainty in conclusions drawn from limited empirical evidence.
    
    \item \textbf{Feedback loops} connect various domains through iterative refinement. Epistemic interpretations inform empirical investigation strategies, while empirical observations refine epistemic models. These bidirectional connections allow the system to navigate adaptively between data collection and knowledge creation based on emerging patterns and insights.
\end{itemize}

These mechanisms facilitate knowledge transformation across different domains of expertise and practice \cite{carlile2004transferring,akkerman2011boundary}.

These boundary-crossing mechanisms are crucial for intelligent microscopy that effectively bridges the epistemic-empirical divide. They allow systems to make informed decisions about when to collect additional data to enhance understanding, when current observations need reinterpretation, or when existing models require refinement based on new evidence.

\subsection{Technical Requirements}
Implementing this architecture presents significant technical challenges that must be addressed for successful deployment:
\begin{itemize}
    \item \textbf{Computational infrastructure} must support both data-intensive processing and knowledge-intensive reasoning. This necessitates  the integration of high-performance computing for image analysis, knowledge representation systems for biological modeling, and real-time processing capabilities for adaptive experimentation. Edge computing approaches that distribute processing between local and cloud resources may be especially valuable in managing the computational demands of smart microscopy.
    \item \textbf{Data standardization approaches} must preserve both empirical details and epistemic context. While existing formats like OME-TIFF can address technical metadata, next-generation standards must also capture experimental context, analytical decisions, and interpretive frameworks. This comprehensive data representation is essential for meaningful sharing of microscopy data and for systems that can reason across datasets and experiments. These approaches align with FAIR (Findable, Accessible, Interoperable, Reusable) principles in bioimage analysis, which similarly emphasize the necessity for contextual metadata, standardized formats, and machine-readable annotations to enable knowledge integration across different imaging studies.
    \item \textbf{System interoperability considerations} must address the integration of existing microscopy platforms, analysis tools, and laboratory information systems. Smart microscopy cannot require complete replacement of research infrastructure but must augment and enhance existing capabilities through modular components and standardized interfaces. This can be achieved by identifying key modules for abstraction across different system components. For instance, device control can be implemented through several approaches: direct hardware access via standard serial communication, device-specific drivers, manufacturer APIs, open-source control frameworks like $\mu$Manager or EPICS, or even GUI automation when other options are unavailable. The critical design requirement is that higher-level planning modules can set or read any device parameter regardless of the underlying implementation, thus creating a unified control interface across heterogeneous microscopy systems.
    \item \textbf{User interface design} must facilitate meaningful human-AI collaboration across technical and conceptual levels. Interfaces should enable researchers to engage with the system at appropriate levels of abstraction—from technical discussions about imaging parameters to conceptual dialogues about interpretation and future experiments. This requires both visualization approaches that effectively convey complex information and interaction models that enhance collaborative scientific reasoning. Currently, no interface fully addresses the human-AI collaboration challenge in microscopy. Even sophisticated platforms like Arkitekt provide advanced workflow capabilities; however they do not specifically cater to human-AI collaborative knowledge creation, although they incorporate foundational elements for knowledge integration. New interface paradigms must be explored and developed to support multi-level interactions between researchers and intelligent microscopy systems, with particular attention to fostering dialogue across the empirical-epistemic boundary\cite{heer2012interactive}.
\end{itemize}

These technical requirements underscore the interdisciplinary nature of smart microscopy development, requiring expertise in optical engineering, computer science, user interface design, and biological research. Addressing these challenges will require collaborative efforts across these domains and the establishment of standards and platforms that facilitate integration across traditionally separate areas of technology.

\subsection{Integration Challenges}
Beyond technical implementation, several challenges in integration must be addressed for the successful adoption of intelligent microscopy in research environments:

\begin{itemize}
    \item \textbf{Laboratory workflow integration} must consider both empirical requirements and epistemic practices of cellular investigation \cite{grudin1994computer}. Systems must align with existing research processes—sample preparation protocols, experimental designs, and analysis pipelines—while enhancing rather than disrupting established scientific workflows.
    \item \textbf{Researcher interaction models} must support appropriate trust calibration and collaborative decision-making \cite{jasanoff2004idiom}. Researchers need to understand system capabilities and limitations, recognize when to rely on automated processes versus when to intervene, and maintain adequate oversight of both data collection and interpretation.
    \item \textbf{Knowledge management approaches} maintain connections between data and meaning across research projects and time. This includes tracking the provenance of both empirical observations and epistemic interpretations, managing evolving understanding, and supporting knowledge transfer between research contexts. Systems should allow researchers to dynamically and explicitly specify contexts or import existing contexts into the knowledge base. These contexts establish the namespace or scope for objects and their interactions within a project-level study, defining the relevant biological entities, experimental conditions, and theoretical frameworks. Context specification can also drive downstream experimental design, including choices of cell lines, visualization methods, small molecule treatments, genes of interest, and other experimental parameters. This contextualization enables intelligent microscopy systems to situate observations appropriately and supports knowledge accumulation across related investigations.
    
    \item \textbf{Ethical considerations} must address issues of data ownership, attribution of scientific insights, and responsible use of automated experimentation. As systems become more active participants in scientific inquiry, clear frameworks must be established for determining appropriate roles, responsibilities, and governance of intelligent microscopy applications. This represents largely unexplored territory in scientific instrumentation. As microscopy systems evolve from passive tools to active partners in knowledge creation, new questions emerge about intellectual contribution, authorship, research integrity, and accountability. These issues require thoughtful consideration of how automated systems can complement human scientific judgment while maintaining appropriate human oversight of the scientific process \cite{sismondo2010introduction}.
\end{itemize}

Addressing these integration challenges requires not only technical solutions but also careful consideration of the social, institutional, and ethical dimensions of scientific research. The successful implementation of smart microscopy will depend on approaches that enhance the collaborative nature of scientific inquiry while respecting the values and practices that underpin research communities.

\section{Conclusion}
\subsection{The Future of Scientific Partnership}
The theoretical framework we have presented reconceptualizes smart microscopy as more than just an advanced imaging tool—it positions these systems as collaborative partners in the scientific process of creating meaning from microscopic observation \cite{king2004functional}. This represents not only a technological advance but also a fundamental shift in our understanding of the relationship between scientific instruments and the process of discovery.

The unique epistemological challenges of cellular investigation—its episodic nature, multi-scale complexity, and the sophisticated cognitive processes through which researchers construct coherent understanding from necessarily fragmented observations—demand systems that can meaningfully participate in navigating the boundary between empirical observation and epistemic understanding. By bridging the divide between what can be directly observed and what must be understood, smart microscopy has the potential to transform how cellular investigations unfold and how biological knowledge emerges \cite{nielsen2020reinventing, leonelli2019data_}.

Our design principles—epistemic-empirical awareness, hierarchical context integration, evolution from detection to perception, adaptive measurement frameworks, narrative synthesis capabilities, and cross-contextual reasoning—serve as a foundation for developing systems that engage with both the technical complexities of microscopy and the conceptual challenges of biological interpretation. The proposed multi-agent architecture provides a practical approach to implementing these principles while addressing the boundary-crossing challenges that have traditionally separated data collection from knowledge creation.

\subsection{Redefining Scientific Instrumentation}
This framework has implications that extend beyond microscopy itself. It suggests a broader reconceptualization of scientific instrumentation, shifting from empirical tools to epistemic partners—from devices that collect data to systems that participate in the creation of understanding. While microscopy presents particularly pronounced epistemic-empirical challenges, all scientific investigations navigate this fundamental divide between observation and understanding \cite{ihde1991instrumental}.

The principles we have outlined may inform the development of intelligent systems across scientific disciplines—from genomics and proteomics to materials science and environmental monitoring \cite{Baird2004ThingKA, Hacking1992-HACTSO-2}. In each domain, the primary challenge lies not merely in automating data collection or standardizing analysis but in creating systems that can meaningfully engage with the process through which empirical observations are transformed into scientific knowledge.

This reconceptualization aligns with emerging views on human-AI collaboration that emphasize complementary intelligence rather than replacement. Smart microscopy systems should enhance distinctly human capabilities—creative exploration, contextual integration, analogical reasoning—rather than simply automating routine tasks. The aim is to amplify human insight instead of substituting human judgment.

\subsection{Metrics of Success: From Efficiency to Understanding}
As we develop smart microscopy systems, we must reconsider how we evaluate their success. Traditional metrics that focus on technical capabilities—resolution, throughput, computational efficiency—remain important but are insufficient \cite{Kuhn_1977}. A truly intelligent microscopy system must be assessed by its contribution to scientific understanding, not merely its operational parameters.

Meaningful evaluation metrics might include:
\begin{enumerate}
    \item \textbf{Discovery acceleration}---does the system enable researchers to identify significant patterns and develop insights more rapidly than traditional approaches?
    \item \textbf{Integration capability}---can the system effectively connect observations across different experimental contexts, modalities, and scales to create a coherent understanding?
    \item \textbf{Adaptive relevance}---does the system dynamically adjust its operation to focus on scientifically relevant phenomena rather than exhaustively collecting predetermined data?
    \item \textbf{Insight suggestion}---can the system propose interpretations, hypotheses, or follow-up experiments that researchers find valuable and would not have immediately considered?
    \item \textbf{Knowledge representation}---does the system create descriptions and visualizations that effectively communicate biological significance and support scientific reasoning?
\end{enumerate}

These approaches to evaluation recognize the social and epistemic dimensions of scientific progress \cite{longino1990, kitcher2001}.

These metrics shift evaluation from technical performance to scientific partnership---from how efficiently systems execute predefined tasks to how meaningfully they contribute to the process of scientific inquiry.

\subsection{The Evolving Relationship Between Researchers and Intelligent Systems}
The development of smart microscopy invites reflection on the evolving relationship between researchers and intelligent systems in scientific investigation \cite{heer2019}. Rather than viewing this relationship through the lens of automation or replacement, we suggest understanding it as a collaborative partnership that navigates the boundaries of the observable and the knowable.

In this partnership, human researchers maintain critical roles that no technological system can replace—setting research priorities, evaluating the significance of findings, connecting observations to broader theoretical frameworks, and ultimately determining what constitutes meaningful scientific progress \cite{clark2003, Collins2007}. Intelligent systems complement these capabilities by enhancing perception, managing complexity, identifying patterns, and bridging episodic gaps in observation.

This collaborative model acknowledges the remarkable capabilities of human scientific cognition as well as its inherent limitations when confronted with the complexity and scale of cellular phenomena. It recognizes that progress does not stem from excluding humans from the scientific process, but rather from developing technologies that extend and enhance distinctly human abilities to derive meaning from observation.

\subsection{Final Reflections}
The framework we have presented represents an ambitious vision for the future of microscopy—one that requires advances not only in optical engineering and computational methods but also in our conceptual understanding of scientific instrumentation and discovery \cite{Snow_Collini_2012}. Realizing this vision will demand interdisciplinary collaboration across optics, computer science, artificial intelligence, user interface design, and cellular biology \cite{Galison1996, Nowotny2003ReThinkingSK}.

Yet the potential rewards are profound. By creating microscopy systems that can meaningfully participate in the navigation between empirical observation and epistemic understanding, we can dramatically enhance our collective capacity to investigate cellular phenomena, accelerate biological discovery, and deepen our understanding of the fundamental processes that underlie life itself. In this way, smart microscopy becomes not merely an advanced tool but a transformative approach to scientific inquiry—one that respects the inherent complexity of both cellular systems and the human process of deriving meaning from our observations.

\section*{Acknowledgements}
We wish to thank Mario Krenn for the initial inspiration to work on automation of science. We are grateful to Karl Johansson, Johannes Kumra Ahnlide, and Oscar André for lively discussions during the days of winter. Special thanks to Svetlana Sukhotskaya for reading the earlier drafts of this manuscript. This work was supported in part by the Wenner-Gren Foundation through their postdoctoral funding program.

\bibliographystyle{unsrt}
\bibliography{references}

\end{document}